\documentclass[10pt,twocolumn,letterpaper]{article}

\usepackage{cvpr}
\usepackage{times}
\usepackage{epsfig}
\usepackage{graphicx}
\usepackage{amsmath}
\usepackage{amssymb}
\usepackage{subfig}
\usepackage{booktabs}
\usepackage{bm}
\usepackage{listings}
\usepackage{color}

\definecolor{dkgreen}{rgb}{0,0.6,0}
\definecolor{gray}{rgb}{0.5,0.5,0.5}
\definecolor{mauve}{rgb}{0.58,0,0.82}

\newcommand{\Zongxin}[1]{#1}

\usepackage[breaklinks=true,bookmarks=false]{hyperref}

\cvprfinalcopy 

\def\cvprPaperID{5772} 

\begin{document}

\title{Gated Channel Transformation for Visual Recognition}

\renewcommand{\thefootnote}{*}
\author{Zongxin Yang$^{1,2}$, Linchao Zhu$^{2}$, Yu Wu$^{1,2}$, and Yi Yang$^{2*}$ \\
$^{1}$ Baidu Research $^{2}$ ReLER, University of Technology Sydney \\
{\tt\small \{zongxin.yang,yu.wu-3\}@student.uts.edu.au, \{linchao.zhu,yi.yang\}@uts.edu.au}
}

\maketitle
\thispagestyle{empty}

\footnotetext{This work was done when Zongxin Yang and Yu Wu interned at Baidu Research. Yi Yang is the corresponding author.}

\begin{abstract}
In this work, we propose a generally applicable transformation unit for visual recognition with deep convolutional neural networks. This transformation explicitly models channel relationships with explainable control variables. These variables determine the neuron behaviors of competition or cooperation, and they are jointly optimized with the convolutional weight towards more accurate recognition.
In Squeeze-and-Excitation (SE) Networks, the channel relationships are implicitly learned by fully connected layers, and the SE block is integrated at the block-level. We instead introduce a channel normalization layer to reduce the number of parameters and computational complexity. This lightweight layer incorporates a simple $\ell_2$ normalization, enabling our transformation unit applicable to operator-level without much increase of additional parameters. Extensive experiments demonstrate the effectiveness of our unit with clear margins on many vision tasks, \ie, image classification on ImageNet, object detection and instance segmentation on COCO, video classification on Kinetics.

\end{abstract}

\section{Introduction}
Convolutional Neural Networks (CNNs) have proven to be critical and robust in visual recognition tasks, such as image classification~\cite{sota_class}, detection~\cite{sota_detec}, and segmentation~\cite{sota_detec}. Notably, a single convolutional layer operates only on a neighboring local context of each spatial position of a feature map, which could possibly lead to local ambiguities~\cite{torralba2003contextual}. To relieve this problem, VGGNets~\cite{vgg} were proposed to construct deep CNNs, using a series of convolutional layers with non-linear activation functions and downsampling operators to cover a large extent of context. Moreover, \cite{resnet} introduced a residual connection to help CNNs benefit from deeper architectures further.

Apart from improving the depth of CNNs, another branch of methods focuses on augmenting the convolutional layer with modules that directly operate on context across large neighborhoods. Squeeze-and-Excitation Networks (SE-Nets)~\cite{senet} leveraged globally embedding information to model channel relationships and modulate feature maps on the channel-wise level. Moreover, its following method, GE-Nets~\cite{genet}, used largely neighboring embedding instead. These modules can be conveniently assembled into modern networks, such as ResNets~\cite{resnet} and Inception~\cite{inception} networks, to improve the representational ability of networks.

\begin{figure}[!t]
\centering

\includegraphics[width=0.98\linewidth]{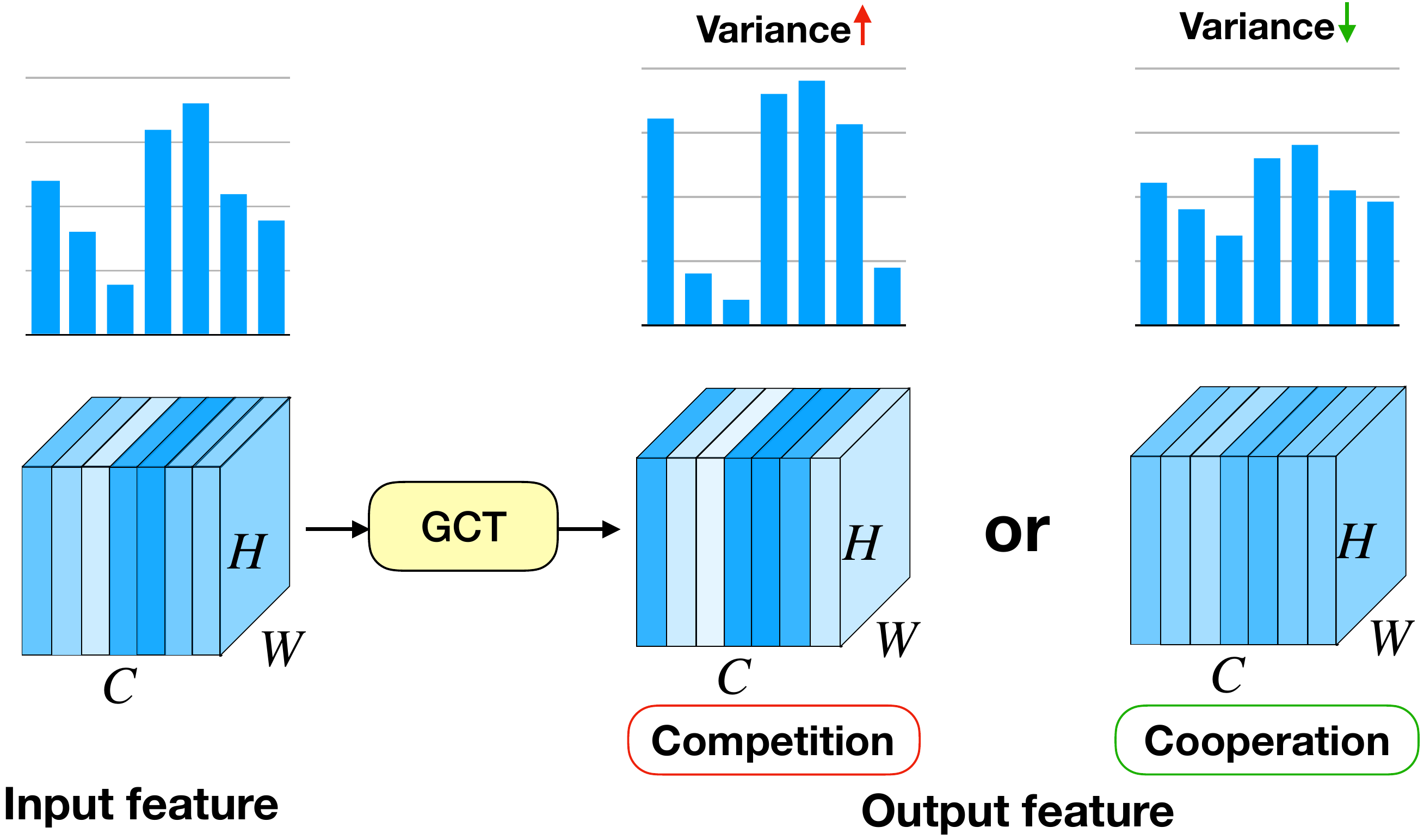}
\caption{An illustration of the behavior of GCT. Combining normalization methods and gating mechanisms, GCT can create the channel relations of both competition (increasing the variance of channel activation) and cooperation (decreasing the variance of channel activation).
}
\label{fig:channel_relation}\vspace{-3mm}
\end{figure}

However, the SE module uses two fully connected (\textit{FC}) layers to process channel-wise embeddings, which leads to two problems. First, the number of SE modules to be applied in CNNs is limited. In~\cite{senet}, SE module was applied at the block-level, \ie, a single SE module is utilized per Res-block~\cite{resnet} or Inception-block~\cite{inceptionv3}. The dimension of the \textit{FC} layer is decreased to save the computational cost further. However, the designed \textit{FC} layers still hinder the wide deployment of SE modules across all layers.
Second, due to the complexity of the parameters in \textit{FC} (or convolutional layer in GE), it is difficult to analyze the interactions among the channels at different layers.
The channel relationships learned by convolution and \textit{FC} operations are inherently implicit~\cite{senet}, resulting in agnostic behaviors of the neuron outputs.

In this paper, we propose a Gated Channel Transformation (GCT) for efficient and accurate contextual information modeling. First, we use a normalization component, instead of \textit{FC}, to model channel relations. Normalization methods, \eg, Local Response Normalization (LRN)~\cite{alexnet}, can create competitions among different neurons (or channels) in neural networks.
Batch normalization~\cite{bn} and its variants can smooth gradient and have been widely used in accelerating CNNs training process.
We leverage a simple $\ell_2$ normalization for modeling channel relationship, which is more stable and computationally efficient comparing to \textit{FC} layers.
Second, we carefully design the trainable architecture of GCT based on the normalization component and introduce a few channel-wise parameters to control the behavior of the gated adaptation of feature channels. Compared to the large number of parameters in \textit{FC}, our designed parameters are much more lightweight. Moreover, the channel-wise gating weight parameter is convenient for channel relationship analysis and helps understand the effect of GCT modules across a whole CNN network.

Following SE, our GCT employs gating mechanisms to adapt the channel relationships. However, the $Sigmoid$ activation of SE is easy to cause vanishing gradient in training, when the $Sigmoid$ value is close to either $0$ or $1$. Hence, we introduce the residual connection~\cite{resnet} into the gating adaptation by using a $1+tanh(x)$ gate activation, which gives GCT an ability to model identity mapping and makes the training process more stable.
Combining normalization methods and gating mechanisms, GCT can create the channel relations of both competition and cooperation, as shown in Fig.~\ref{fig:channel_relation}. 
According to our visualization analysis (Sec.~\ref{sec:analysis}), GCT prefers to encourage cooperation in shallower layers, but competition is enhanced in deeper layers. 
Generally, the shallow layers learn low-level attributes to capture general characteristics like textures. In deeper layers, the high-level features are more discriminative and task-related.

Our experiments show that GCT is a simple and effective architecture for modeling relationships among channels. It significantly improves the generalization capability of deep convolutional networks across visual recognition tasks and datasets.


\section{Related Work}
\noindent\textbf{Gating and attention mechanisms.}
 Gating mechanisms have been successfully deployed in some recurrent neural network architectures. Long Short-Term Memory (LSTM)~\cite{lstm} introduced an input gate, output gate and forget gate, which are used to regulate the flow of information into and out of the module. 
 Based on gating mechanisms, some attention methods focus on forcing computational resources towards the most informative components of features~\cite{attention3, attention4}. 
The attention mechanism has achieved promising improvements across many tasks including sequence learning~\cite{sequence_learning1}, lip reading~\cite{lip_reading}, image captioning~\cite{image_caption1}, localization and understanding in images~\cite{localiztion, understanding}.

 Recent works introduce the attention mechanism into convolutional networks (\eg,~\cite{attention_conv1, attention_conv2}). 
A non-recurrent approach to combine gating mechanisms with convolutional networks achieves promising performance in the language task, which was always studied based on recurrent networks before~\cite{attention_conv2}.
 Following these studies, SE-Nets~\cite{senet} and its following work GE-Nets~\cite{genet} introduced a lightweight gating mechanism that focuses on enhancing the representational power of the convolutional network by modeling channel-wise relationship. Compared to the SE module, our GCT also pays attention to the cross-channel relationship but can achieve better performance gains with less computation and parameters.

\noindent\textbf{Normalization layers.}
In recent years, normalization layers have been widely used in deep networks to create competition between neurons~\cite{alexnet} and produce smoother optimization surfaces~\cite{bn}. Local Response Normalization (LRN)~\cite{alexnet} computes the statistics in a small neighborhood among channels for each pixel.
Batch Normalization (BN)~\cite{bn} utilizes global spatial information along the batch dimension and suggests to be deployed for all layers. 
Layer Normalization (LN)~\cite{ln} computes along the channel dimension instead of the batch dimension. Group Normalization (GN)~\cite{gn} differently divides the channels into groups and computes within each group the mean and variance for normalization. Similar to LRN, GN and LN, our GCT also utilizes channel-related information with normalization structure.

\noindent\textbf{Deep architectures.}
VGGNets~\cite{vgg} and Inception networks~\cite{inception} demonstrated that it was significant to improve the quality of representation by increasing the depth of a network. ResNets~\cite{resnet} utilized shortcut connections to identity-based skip connections, and proved that it was highly effective to build considerably deeper and stronger networks with them. 
Some other researchers focused on improving the representation ability of the computational elements contained within a network~\cite{inceptionv3}. The more diverse composition of operators within a computational element can be constructed with multi-branch convolutions or pooling layers. Other than this, grouped convolutions have proven to be a practical method to increase the cardinality of learned transformations~\cite{resnext}.

Based on the success of CNNs for image tasks, 3D convolutions~\cite{c3d1,c3d2} (C3D) are introduced for video classification task. In addition to C3D, Non-local neural networks~\cite{nlnet} (NL-Nets) design a non-local operation for capturing long-range, non-local dependency, which significantly improves the accuracy of video classification.

\begin{figure*}[!t]
\centering

\includegraphics[width=0.8\linewidth]{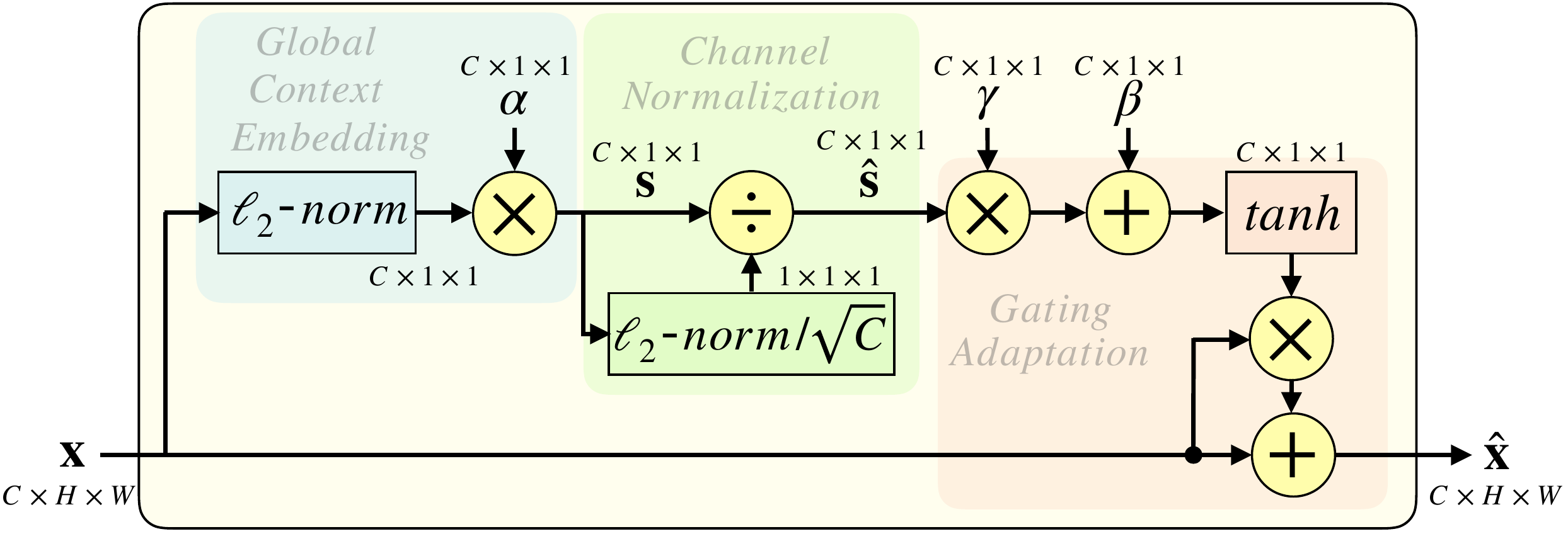}

\caption{An \textbf{overview} of the structure of Gated Channel Transformation (GCT).
The embedding weight, $\bm{\alpha}$, is responsible for controlling the weight of each channel before the channel normalization. And the gating weight and bias, $\bm{\gamma}$ and $\bm{\beta}$, are responsible for adjusting the scale of the input feature $\bm{x}$ channel-wisely.
}
\label{fig:model}
\end{figure*}

We build our GCT on some of these deep architectures. All the networks with GCT achieve promising performance improvements, but the growth of computational complexity is negligible.


\section{Gated Channel Transformation}
We propose a Gated Channel Transformation for highly efficient, channel-wise, contextual information modeling. GCT employs a normalization method to create competition or cooperation relationships among channels. Notably, the normalization operation is parameter-free. To make GCT learnable, we design a global context embedding operator, which embeds the global context and controls the weight fo each channel before the normalization and a gating adaptation operator, which adjusts the input feature channel-wisely based on the output of the normalization. The channel-wise trainable parameters are light-weight yet effective and make GCT convenient to be extensively deployed while occupying a small number of parameters. Besides, the parameters of the gating adaptation operator are easy and intuitive to be visualized for explaining the behavior of GCT. In summary, we carefully design the highly light-weight, explainable, but effective architecture of GCT based on the normalization operation for modeling channel relationships.

Let $\mathbf{x}\in\mathbb{R}^{C\times H\times W}$ be an activation feature in a convolutional network, where $H$ and $W$ are the spatial height and width, and $C$ is the number of channels. In general, GCT performs the following transformation:
\begin{equation} \label{equ:overview}
    \hat{\mathbf{x}}=F(\mathbf{x}|\bm{\alpha},\bm{\gamma},\bm{\beta}), \bm{\alpha},\bm{\gamma},\bm{\beta}\in\mathbb{R}^{C}.
\end{equation}
Here $\bm{\alpha}$, $\bm{\gamma}$ and $\bm{\beta}$ are trainable parameters.
Embedding weight $\bm{\alpha}$ is responsible for adapting the embedding outputs.
The gating weight $\bm{\gamma}$ and bias $\bm{\beta}$ control the activation of the gate. They determine the behavior of GCT in each channel.
Notably, the parameter complexity of GCT is $O(C)$, which is smaller than the SE module ($O(C^2)$)~\cite{senet}. In SE-Net, two \textit{FC} layers are leveraged, which have the parameter complexity of $O(C^2)$.

An illustration of the structure of GCT is shown in Fig.~\ref{fig:model}. Let $\mathbf{x}=[x_1, x_2,...,x_C], x_c=[x_c^{i,j}]_{H\times W}\in\mathbb{R}^{H\times W}, c\in\{ 1,2,...,C\}$, where $x_c$ is corresponding to each channel of $\mathbf{x}$. The detailed transformation consists of following parts.

\subsection{Global Context Embedding} 
The information with a large receptive field is useful to avoid local ambiguities~\cite{torralba2003contextual,genet} caused by the information with a small receptive field (\eg, a convolutional layer). Hence, we firstly design a
global context embedding module to aggregate global context information in each channel. The module can exploit global contextual information outside the small receptive fields of convolutional layers. Given the embedding weight $\bm{\alpha}=[\alpha_1,...,\alpha_C]$, the module is defined as:
\begin{equation} \label{adaption1}
    s_c =\alpha_c||x_{c}||_2=\alpha_c\{[\sum\limits_{i=1}^{H}\sum\limits_{j=1}^{W} (x_{c}^{i,j})^2]+\epsilon\}^\frac{1}{2},
\end{equation}
where $\epsilon$ is a small constant to avoid the problem of derivation at the zero point. 
Different from SE, GCT does not use global average pooling (GAP) to aggregate channel context. GAP might fail in some extreme cases. For example, if SE is deployed after the Instance Normalization~\cite{in} layer that is popular in style transfer task, the output of GAP will be constant for any inputs since IN fixes the mean of each channel of features. To avoid this problem, we choose $\ell_p$-norm instead.
It is worth noting that GCT is robust with different $\ell_p$-norms.
In Sec.~\ref{sec:embedding}, we compare the performance of some popular $\ell_p$-norms and choose the best one, $\ell_2$-norm, to be our default setting. Notably, the performance of $\ell_1$-norm is very close to $\ell_2$-norm, but $\ell_1$-norm can be equivalently replaced by GAP when the input of GCT is consistently non-negative (for example, after ReLU activation in our default setting). In this case, $\ell_1$-norm is more computationally efficient as shown in Table~\ref{tab:clock}.

Besides, we use trainable parameters, $\alpha_c$, to control the weight of each channel because different channels should have different significance. Especially if $\alpha_c$ is close to $0$, the channel $c$ will not be involved in the channel normalization. In other words, the gating weight, $\bm{\alpha}$, make GCT capable of learning the situation that one channel is individual to other channels.

\subsection{Channel Normalization}
Normalization methods can create competition relationship between neurons (or channels)~\cite{alexnet} with lightweight computing resource and a stable training performance (\eg, \cite{bn}).
Similar to LRN, we use a $\ell_2$ normalization to operate across channels, namely channel normalization. Let $\mathbf{s}=[s_1,...,s_C]$, the formula of channel normalization is:

\begin{equation} \label{adaption2}
    \hat{s}_c =\frac{\sqrt{C} s_c}{||\mathbf{s}||_2}=\frac{\sqrt{C} s_c}{[(\sum\limits_{c=1}^{C} s_{c}^{2})+\epsilon]^\frac{1}{2}},
\end{equation}
where $\epsilon$ is a small constant.
The scalar $\sqrt{C}$ is used to normalize the scale of $\hat{s}_c$,  avoiding a too small scale of $\hat{s}_c$ when $C$ is large. 
Compared to the \textit{FC} layers used by SE, our channel normalization has less computational complexity ($O(C)$) compared to the \textit{FC} layers ($O(C^2)$).

\subsection{Gating Adaptation}
We employ a gating mechanism, namely gating adaptation, to adapt the original feature. By introducing the gating mechanism, our GCT can facilitate both competition and cooperation during the training process.
Let the gating weight $\bm{\gamma}=[\gamma_1,...,\gamma_C]$ and the gating biases $\bm{\beta}=[\beta_1,...,\beta_C]$, we design the following gating function:
\begin{equation} \label{adaption3}
    \hat{x}_c =x_c[1+\text{tanh}(\gamma_c\hat{s}_c+\beta_c)].
\end{equation} 
The scale of each original channel $x_c$ will be adapted by its corresponding gate, \ie, $1+\text{tanh}(\gamma_c \hat{s}_c+\beta_c)$. Due to the channel normalization is parameter-free, we design the trainable weight and bias, $\bm{\gamma}$ and $\bm{\beta}$, for learning to control the activation of gate channel-wisely. 
LRN benefits from only the competitions among the neurons~\cite{alexnet}. However, GCT is able to model more types of relationship (\ie, competition and cooperation) among different channels by combining normalization methods and gating mechanisms. 
When the gating weight of one channel ($\gamma_c$) is activated positively, GCT promotes this channel to compete with the others as in LRN. When the gating weight is activated negatively, GCT encourages this channel to cooperate with the others. We analyze these adaptive channel relationships in Sec.\ref{sec:analysis}.

Besides, this gate function allows original features to pass to the next layer when the gating weight and biases are zeros, which is 
\begin{equation}
    \hat{\mathbf{x}}=F(\mathbf{x}|\bm{\alpha},\bm{0},\bm{0})=\mathbf{1}\mathbf{x}=\mathbf{x}.
\end{equation}
The ability to model identity mapping can effectively improve the robustness of the degradation problem in deep networks. ResNets also benefit from this idea. Therefore, we propose to initialize $\bm{\gamma}$ and $\bm{\beta}$ to $0$ in the initialization of GCT layers. By doing this, the initial steps of the training process will be more stable, and the final performance of GCT will be better.

\begin{figure}
    \centering
    \vspace{-2mm}
\begin{lstlisting}
def forward(self, x, epsilon=1e-5):
  # x: input features with shape [N,C,H,W]
  # alpha, gamma, beta: embedding weight, gating weight, gating bias with shape [1,C,1,1]
  embedding = (x.pow(2).sum((2,3), keepdim=True) + epsilon).pow(0.5) * self.alpha
  norm = self.gamma / (embedding.pow(2).mean(dim=1, keepdim=True) + epsilon).pow(0.5)
  gate = 1. + torch.tanh(embedding * norm + self.beta)
  return x * gate
\end{lstlisting}
\vspace{-3mm}
    \caption{An implementation of GCT ($\ell_2$) based on PyTorch}
    \label{fig:code}\vspace{-2mm}
\end{figure}

\subsection{Learning}
\label{app:training_detail}
Similar to modern normalization layers (\eg, BN), we propose to apply GCT for all convolutional layers in deep networks. However, there are many different points nearby one convolutional layer to employ GCT. In deep networks, each convolutional layer always works together with a normalization layer (\eg, BN) and an activation layer (\eg, ReLU~\cite{relu}). For this reason, there are three possible points to deploy the GCT layer, which are before the convolutional layer, before the normalization layer, and after the normalization layer. All these methods are effective, but we find it to be better to employ GCT before the convolutional layer. In Sec.~\ref{sec:appication}, we compare the performance of these three application methods. 

We implement and evaluate our GCT module on some popular deep learning frameworks, including PaddlePaddle~\cite{paddlepaddle}, TensorFlow~\cite{abadi2016tensorflow} and PyTorch~\cite{pytorch}, and we observe similar improvement by introducing GCT. Fig.~\ref{fig:code} shows a simple implementation based on PyTorch.

\begin{figure*}[!t]
\center
\vspace{-2mm}
\subfloat[ResNet-50]{
\label{fig:places.before_conv}
\includegraphics[width=0.32\linewidth]{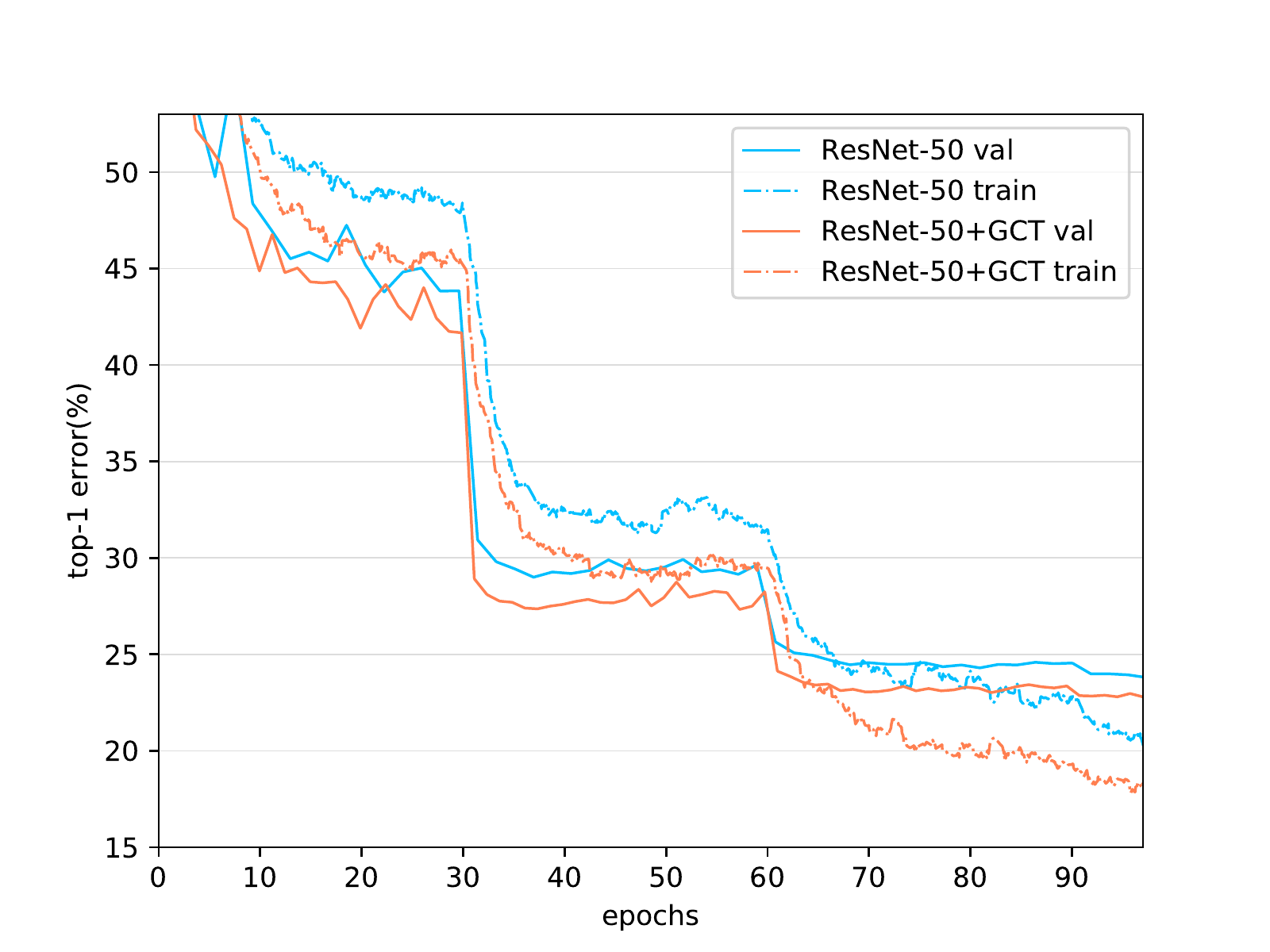}
}
\subfloat[ResNet-101]{
\label{fig:places.before_norm}
\includegraphics[width=0.32\linewidth]{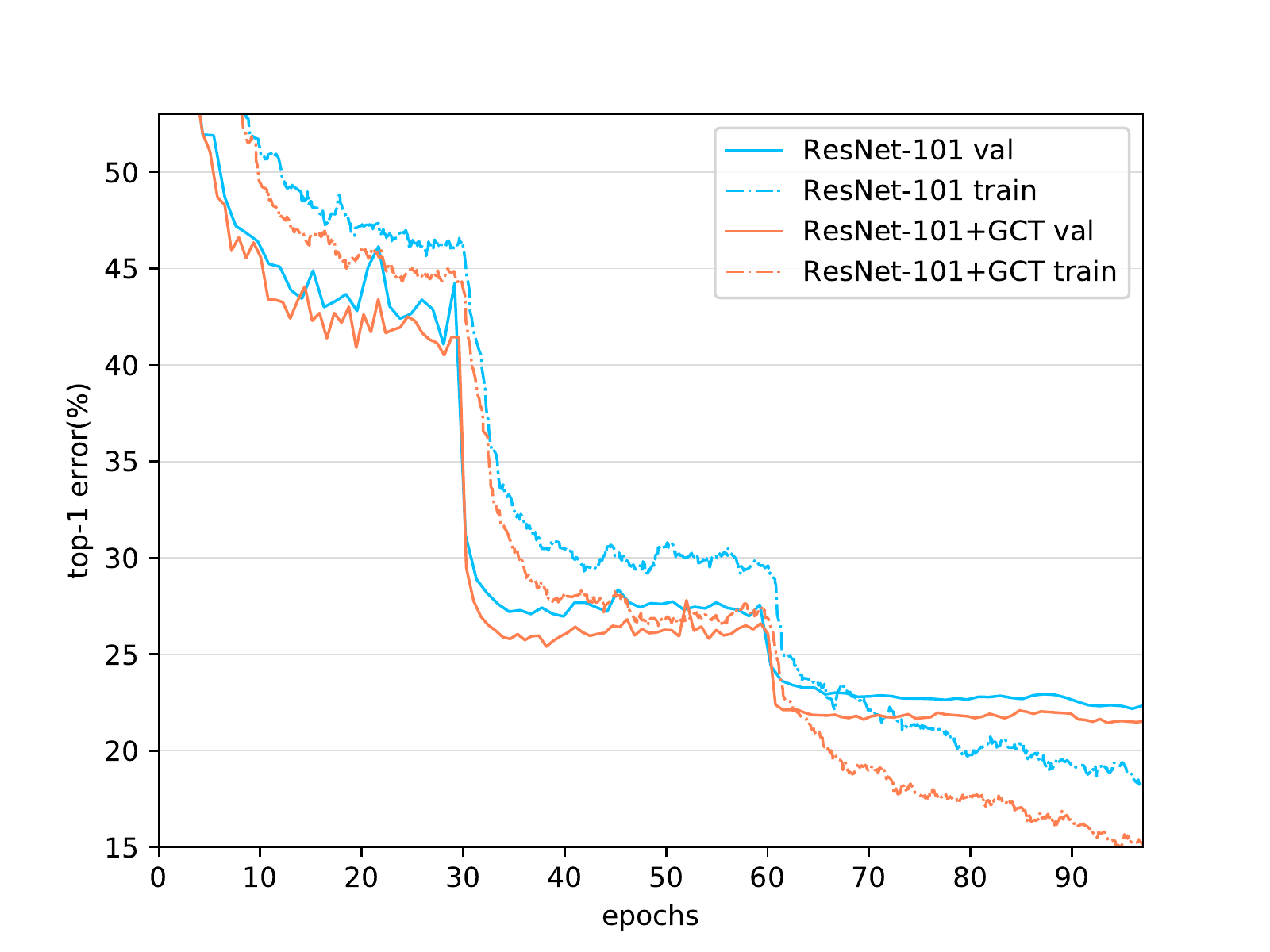}
}
\subfloat[ResNet-152]{
\label{fig:places.after_norm}
\includegraphics[width=0.32\linewidth]{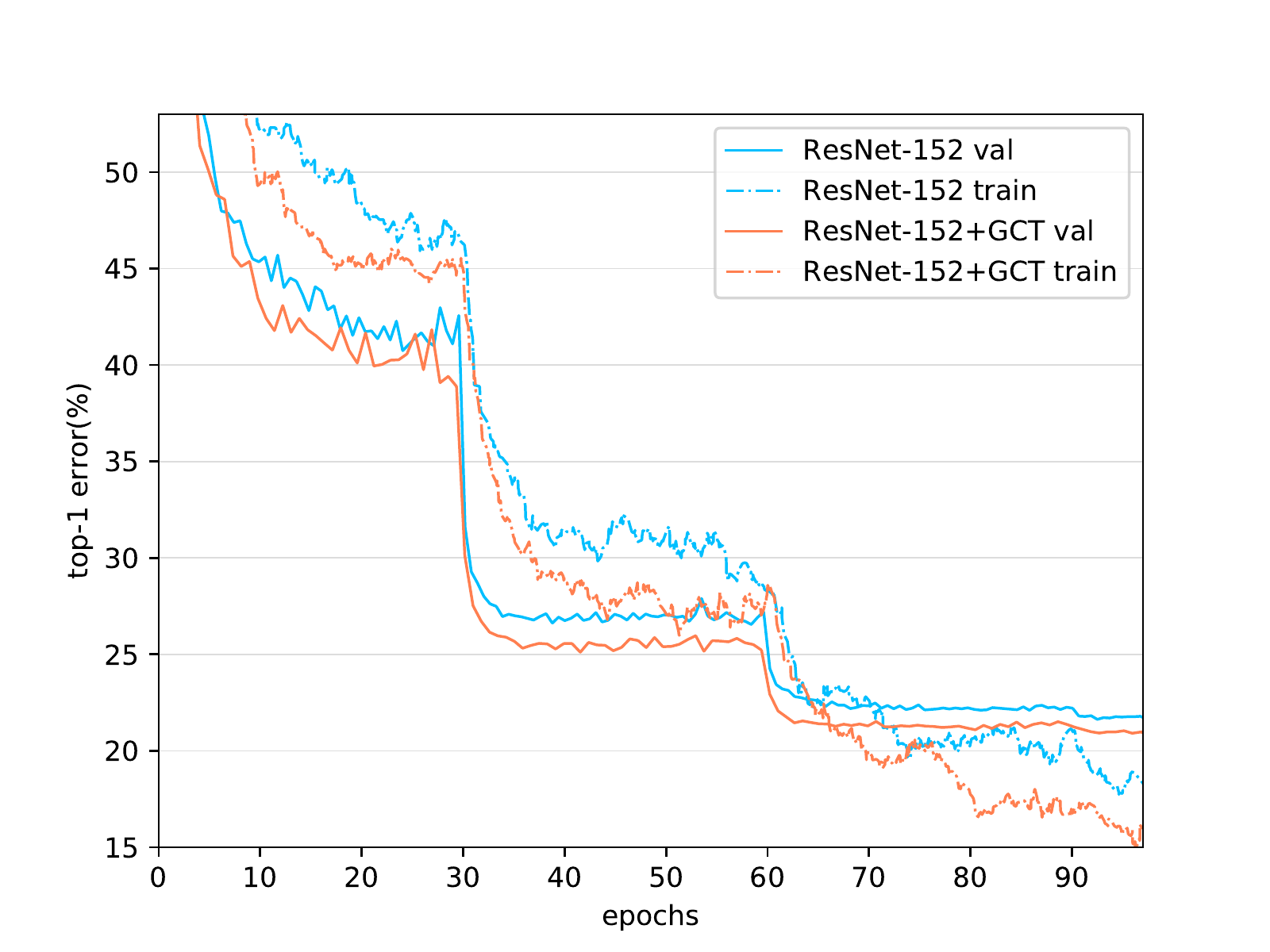}
}
\vspace{-2mm}
\caption{\textbf{Training curve comparisons} for ResNets with different depth on ImageNet.}
\label{fig:curve}
\end{figure*}

In the training process, we propose to use $1$ to initialize $\bm{\alpha}$ and use $0$ to initialize all $\bm{\gamma}$ and $\bm{\beta}$. By doing this, GCT will be initialized as an identity mapping module, which will make the training process more stable. Besides, to avoid the bad influence of unstable gradient on the GCT gate in initial training steps, we propose to use warmup method (to start training with a small learning rate). In all the experiments on ImageNet~\cite{imagenet}, we start training with a learning rate of $0.01$ for $1$ epoch. After the warmup, we go back to the original learning rate schedule. Finally, we propose \textbf{NOT} to apply weight decay on $\bm{\beta}$ parameters, which is possible to reduce the performance of GCT.

\begin{table}[!t]
\footnotesize
\begin{center}

\begin{tabular}{|l|c|c|c|c|}
\hline  
            &    \multicolumn{2}{|c|}{original}  &        \multicolumn{2}{|c|}{\textbf{GCT}}    \\
\hline  
          Network  & top-1 & top-5  & top-1 & top-5    \\
\hline\hline
VGG-16~\cite{vgg}      & 26.2 & 8.3  &  \textbf{25.1$_{(1.1)}$} & \textbf{7.5$_{(0.8)}$} \\
\hline
Inception-v3~\cite{inceptionv3}   & 24.3 & 7.3  & \textbf{23.7$_{(0.6)}$}  & \textbf{7.1$_{(0.2)}$} \\
\hline\hline
ResNeXt-50~\cite{resnext}  & 22.4  & 6.3  &  \textbf{21.7$_{(0.7)}$} & \textbf{6.0$_{(0.3)}$} \\
\hline
ResNet-50~\cite{resnet}   &   23.8    & 7.0 & \textbf{22.7$_{(1.1)}$}  & \textbf{6.3$_{(0.7)}$} \\
\hline
ResNet-101~\cite{resnet}   &  22.2  &  6.2  & \textbf{21.4$_{(0.8)}$}  & \textbf{5.9$_{(0.3)}$} \\
\hline
ResNet-152~\cite{resnet}   & 21.6 &  5.9  & \textbf{20.8$_{(0.8)}$}  & \textbf{5.5$_{(0.4)}$} \\
\hline
ResNet-200$^{*}$~\cite{resnet2}   & 20.7 &  5.2  & \textbf{19.7$_{(1.0)}$}  & \textbf{4.8$_{(0.4)}$} \\
\hline
\end{tabular}
\caption{\textbf{Improvement in error performance (\%) on ImageNet.} The numbers in brackets denote the improvement in performance over the baselines. ResNet-200$^{*}$ means we follow the strategy in~\cite{resnet2} to train this model on $224\times224$ but evaluate on $320\times320$.}\label{expr:final_results}\vspace{-2mm}
\end{center}
\end{table}

\section{Experiments}
We apply GCT for all the convolutional layers in deep networks rather than block-level deployment in SE-Net. In all GCT counterparts, we employ one GCT layer before each convolutional layer. In the Kinetics experiments, we apply GCT at the last two convolutional layers in each Res-Block. More training details are shown in Sec.~\ref{app:training_detail}.

\subsection{ImageNet}
\label{sec:expr_imagenet}

We experiment on the ImageNet 2012 dataset~\cite{imagenet} with $1,000$ classes. We train all the models on the $1.28$M training images and evaluate on the $50,000$ validation images.

\noindent\textbf{Training details.}
In the training process of all the models, the input image is $224\times224$ randomly cropped from a resized image using the same augmentation in \cite{inception}. We use SGD with a mini-batch size of $256$. For ResNet-152 and ResNeXt-50, we use half mini-batch size and double the training steps). The weight decay is $0.0001$, and the momentum is $0.9$. The base learning rate is $0.1$, and we divide it by $10$ every $30$ epochs. All models are trained for $100$ epochs from scratch, using the weight initialization strategy described in~\cite{initialization}. Besides, we start the training process with a learning rate of $0.01$ for $1$ epoch. After the warmup, we go back to the original learning rate schedule. In all comparisons, we evaluate the error on the single  $224\times224$ center crop from an image whose shorter side is $256$.
For ResNet-200~\cite{resnet2}, we evaluate on $320\times320$ following~\cite{resnet2}.

\begin{table*}[!ht]
\footnotesize
\begin{center}
\begin{tabular}{|l|c|c|c|c|c|c|}
\hline  
            &    \multicolumn{2}{|c|}{original}  &    \multicolumn{2}{|c|}{+SE~\cite{senet}} &        \multicolumn{2}{|c|}{\textbf{+GCT (ours)}}    \\
\hline  
          Network  & top-1/5 & G/P  & top-1/5 & G/P & top-1/5 & G/P     \\
\hline\hline
ResNet-50~\cite{resnet} & 23.8/7.0 & 3.879/25.61 &  22.9/6.6  & \textbf{3.893}$^*$/28.14 & \textbf{22.7/6.3} & 3.900/\textbf{25.68} \\
\hline
ResNeXt-50~\cite{resnext} & 22.4/6.3 & 3.795/25.10 &  22.0/6.1  & \textbf{3.809}$^*$/27.63 &  \textbf{21.7/6.0} & 3.821/\textbf{25.19}   \\
\hline
Inception-v3~\cite{inceptionv3} & 24.3/7.3 & 2.847/23.87 & 24.0/7.2  & \textbf{2.851}$^*$/25.53  & \textbf{23.7/7.1} & 2.862/\textbf{23.99} \\
\hline\hline
VGG-16~\cite{vgg} & 26.2/8.3 & 15.497/138.37 &  25.2/7.7   & 15.525/138.60 & \textbf{25.1/7.5} & \textbf{15.516}/\textbf{138.38} \\
\hline
\end{tabular}
\caption{\textbf{Compared to SE on ImageNet.} We evaluate the models of error performance (\%), GFLOPs (G) and parameters (M). G/P means GFLOPs/parameters. $^*$: In the first three networks, SE is only employed in block-level (Res-Block or Inception-Block) as proposed~\cite{senet}, but GCT is applied for all the convolutional layers. This difference makes that SE uses comparable GFLOPs with GCT. In VGG-16, however, SE is employed for all the convolutional layers in the experiments, which is the same as GCT. Under the same setting, GCT outperforms SE on both complexity and performance.}\label{expr:comparison}\vspace{-2mm}
\end{center}
\end{table*}

\begin{table*}[t]
    \centering
    \footnotesize
    \setlength{\tabcolsep}{10pt}
    \begin{tabular}{|l|c|c|c|c|}
        \hline
         & baseline & +SE~\cite{senet} & +GCT ($\ell_1$-norm) & +GCT ($\ell_2$-norm) \\
        \hline
        ResNet-50~\cite{vgg} & $603$ & $\bm{525^*}$ & $484$ & $425$ \\
        \hline
        VGG-16~\cite{resnet} & $313$ & $183$ & $\bm{281}$ & $268$ \\
        \hline
    \end{tabular}
    \caption{\textbf{Inference speed (FPS) comparison}. We compare the speed on ImageNet by using one GTX 1080 Ti GPU. GCT is always applied for all the convolutional layers. $^*$: In ResNet-50, SE is applied for each Res-Block as proposed. But, in VGG-16, SE is applied for all the convolutional layers, which leads to better fairness in the same setting with GCT.}\label{tab:clock}
    
\end{table*}

\noindent\textbf{Integration with deep modern architectures.}
We study the effects of integrating GCT layers with some state-of-the-art backbone architectures, \eg, ResNets and ResNeXts~\cite{resnext}, in which we apply GCT before all the convolutional layers. We report all these results in Table~\ref{expr:final_results}. Compared to original architectures, we observe significant performance improvements by introducing GCT into networks. Particularly, the top-1 error of GCT-ResNet-101 is $\mathbf{21.4}\%$, which is even better than the ResNet-152 baseline ($21.6\%$) with a deeper network and much more parameters. In addition, GCT is able to bring stable improvement in ResNets with different depth ($\mathbf{1.1}\%$ top-1 improvement in ResNet-50, $\mathbf{0.8}\%$ in ResNet-152 and $\mathbf{1.0}\%$ in ResNet-200). Besides, we observe a smooth improvement throughout the training schedule, which is shown in Fig.\ref{fig:curve}.

We also explore the improvement with GCT in \textit{non-residual} networks (\eg, VGG-16~\cite{vgg} and Inception-v3~\cite{inceptionv3}). To stabilize the training process, we employ BN~\cite{bn} layers after every convolutional layer. Similar to the effectiveness in residual architectures, GCT layers bring promising improvements in \textit{non-residual} structures.

\noindent\textbf{Compared to SE.}
Following~\cite{genet}, we conduct experiments on ImageNet to compare SE with GCT. In addition, we make comparison fairly in both residual and \textit{non-residual} networks and the results are reported in Table \ref{expr:comparison}. We follow the methods in \cite{senet} to integrate SE into VGG-16~\cite{vgg}, Inception~\cite{inceptionv3}, ResNet-50~\cite{resnet} and ResNeXt-50~\cite{resnext} and train these models in same training schedule. Compared to SE, GCT always achieves better improvement.

In order to compare computational complexity, we calculate the GFLOPs and the number of parameters. In VGG-16 experiments, SE is employed for all the convolutional layers, which is the same as GCT. Under this fair condition, GCT achieves better performance with less increase in both GFLOPs ($\textbf{0.019}$G \vs $0.028$G) and parameters ($\textbf{0.01}$M \vs $0.23$M). Moreover, GCT is much more efficient than SE in run-time as shown in Table~\ref{tab:clock} ($\mathbf{281}$FPS \vs $183$). In other experiments, SE is employed in block-level (Res-Block or Inception-Block) as proposed~\cite{senet}, which means the number of SE modules is only about $1/3$ of GCT. However, the increase in parameters of GCT is still much less than SE, and the inference speed of GCT is comparable with SE (GCT $484$FPS \vs SE $525$ in ResNet-50). Compared to SE, GCT performs better and is capable of applying for all the convolutional layers while keeping the network efficient.

\subsection{COCO}
Next we evaluate the generalizability on the COCO dataset~\cite{coco}. We train the models on the COCO \textit{train2017} set and evaluate on the COCO \textit{eval2017} set (a.k.a \textit{minival}).

\noindent\textbf{Training details.}
We experiment on the Mask R-CNN baselines~\cite{mask_rcnn} and its GN counterparts~\cite{gn}. All the backbone models are pre-trained on ImageNet using the scale and aspect ratio augmentation in~\cite{inception} and fine-tune on COCO with a batch size of $16$ (2 images/GPU). Besides, all these experiments use the Feature Pyramid Network (FPN)~\cite{fpn}. We also use the same hyperparameters and two training schedules used in \cite{gn}. The short schedule includes $90$K iterations, in which the learning rate is divided by $10$ at $60$K and $80$K iterations. The long schedule increases the iterations to $270$K, in which the learning rate is divided by $10$ at $210$K and $250$K. The base learning rate is $0.02$ in both schedules.

\begin{table}[!t]
\footnotesize
\begin{center}
\begin{tabular}{|l|c|c|c|}
\hline  
  Backbone    &  box head &   box AP & mask AP      \\
\hline\hline
ResNet-50 BN$^{*}$   &  -   &  37.8 & 34.2   \\
\hline
ResNet-50 BN$^{*}$+SE~\cite{gcnet}   &  -   &  38.2$_{(0.4)}$ & 34.7$_{(0.5)}$   \\
\hline
ResNet-50 BN$^{*}$+\textbf{GCT}   &  -   &  \textbf{39.8$_{(2.0)}$} & \textbf{36.0$_{(1.8)}$}   \\
\hline\hline
ResNet-101 BN$^{*}$  &  -  &  40.1 & 36.1   \\
\hline
ResNet-101 BN$^{*}$+\textbf{GCT}  &  -   &  \textbf{42.0$_{(1.9)}$} & \textbf{37.7$_{(1.6)}$}   \\
\hline
\hline \hline 
$^{+}$ResNet-50 BN$^{*}$   &  -   &  38.6 & 34.5   \\
\hline
$^{+}$ResNet-50 GN   &  GN   &  40.8$_{(2.2)}$ & 36.1$_{(1.6)}$   \\
\hline
$^{+}$ResNet-50 BN$^{*}$+\textbf{GCT}   &  GN   & \textbf{41.6$_{(3.0)}$} & \textbf{37.1$_{(2.6)}$}   \\
\hline
$^{+}$ResNet-50 BN$^{*}$+\textbf{GCT}   &  GN+\textbf{GCT}   & \textbf{41.8$_{(3.2)}$} & \textbf{37.3$_{(2.8)}$}   \\
\hline\hline 
$^{+}$ResNet-101 BN$^{*}$   &  -   &  40.9 & 36.4   \\
\hline
$^{+}$ResNet-101 GN   &  GN   &  42.3$_{(1.4)}$ & 37.2$_{(0.8)}$   \\
\hline
$^{+}$ResNet-101 BN$^{*}$+\textbf{GCT}   &  GN   & \textbf{43.1$_{(2.2)}$}  &  \textbf{38.3$_{(1.9)}$}  \\
\hline 
\end{tabular}
\caption{\textbf{Improvement on COCO with Mask R-CNN framework~\cite{mask_rcnn}.} BN$^{*}$ means BN is frozen. $^{+}$ means increasing the training iterations from $90$K to $270$K. When using GN, we follow the original strategy in~\cite{gn}.}\label{expr:final_results_on_coco}\vspace{-2mm}
\end{center}
\end{table}

\begin{figure*}[t!]
\center
\vspace{-2mm}
\subfloat[distribution of $\bm{\gamma}$]{
\label{fig:analysis1}
\includegraphics[width=0.245\linewidth]{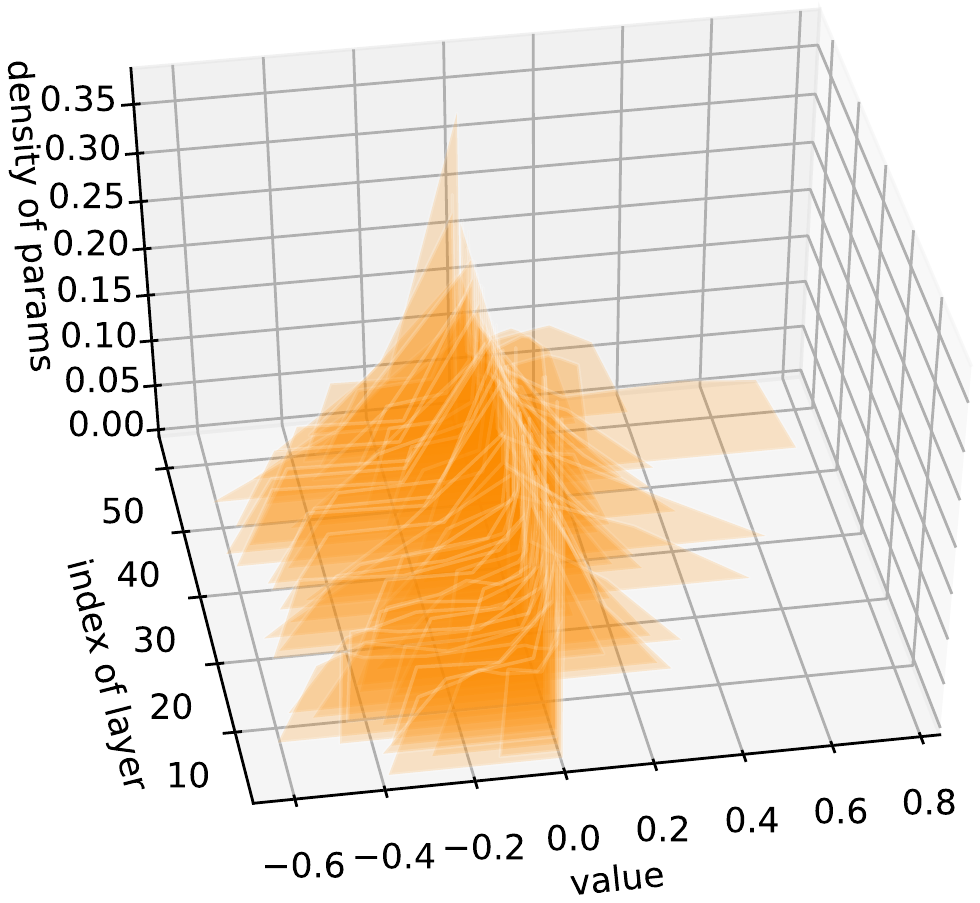}
}
\subfloat[mean (std) of $\bm{\gamma}$]{
\label{fig:analysis2}
\includegraphics[width=0.290\linewidth]{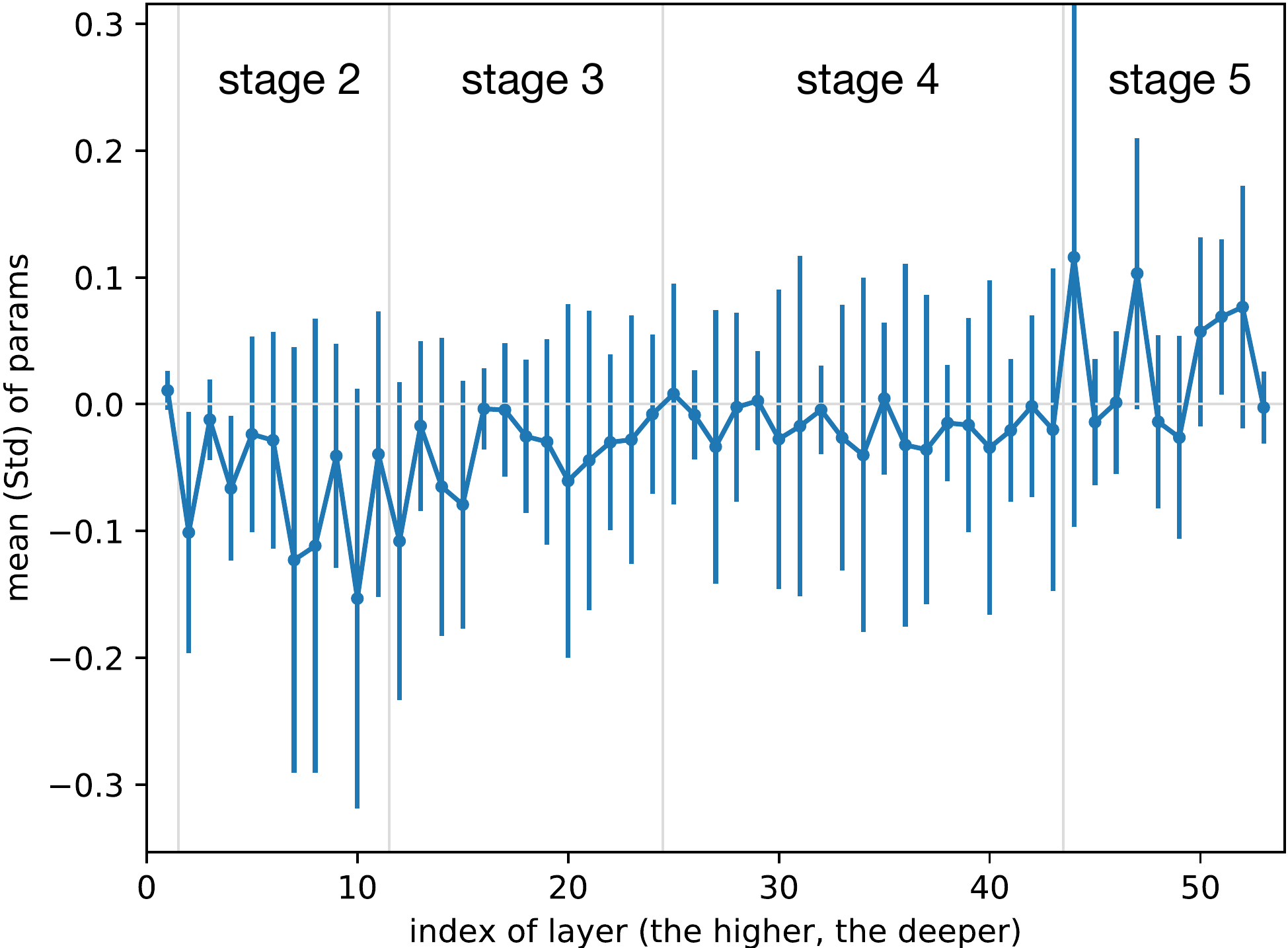}
}
\subfloat[ratio of the variance of output and input]{
\label{fig:analysis3}
\includegraphics[width=0.285\linewidth]{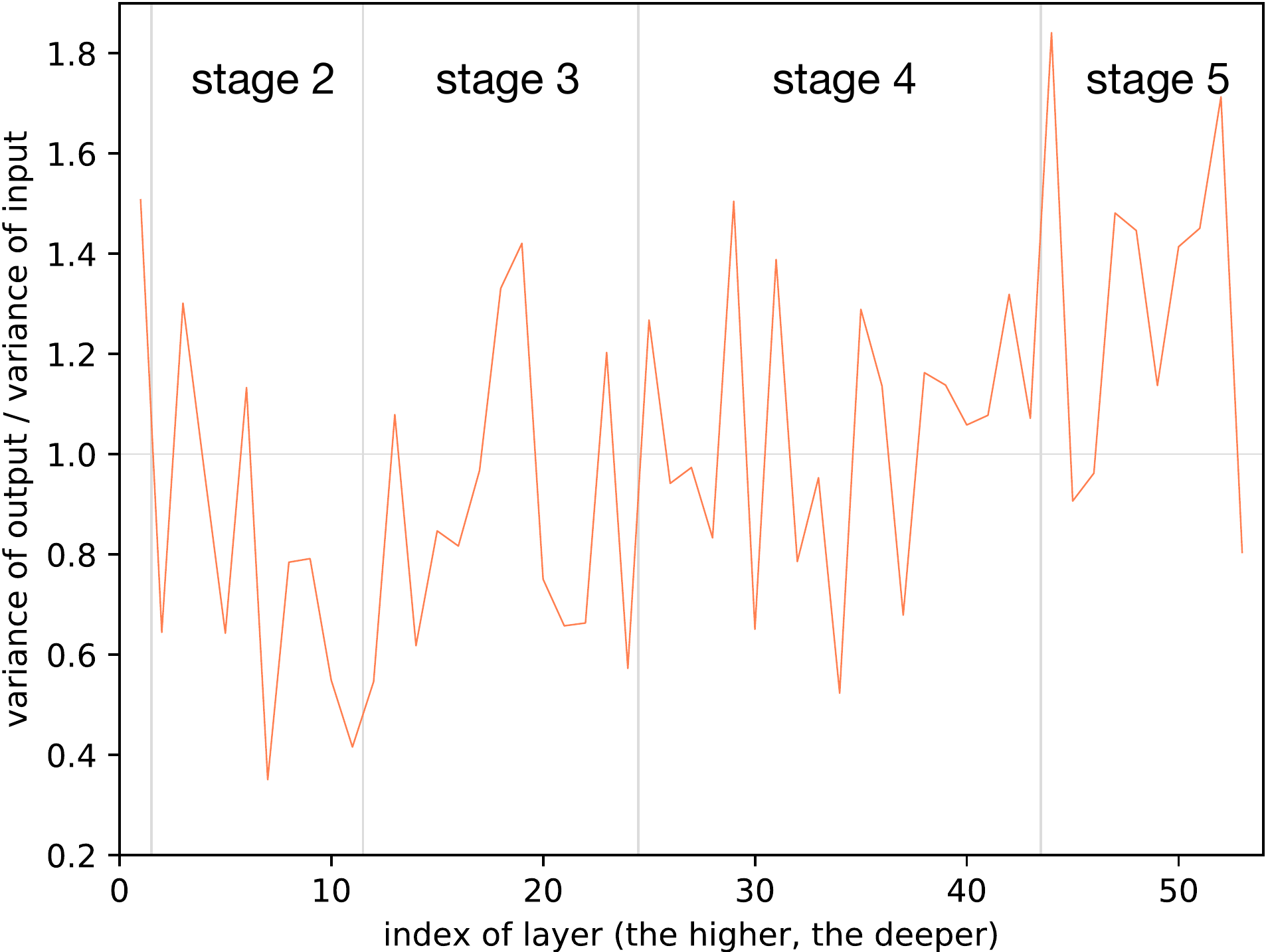}
}

\caption{\textbf{Analysis}. The visulization of parameters of $\bm{\gamma}$ (Fig.(a), (b)), and the ratio of variance of GCT output and input feature (Fig.(c)) in all the GCT layers in ResNet-50 on ImageNet.
}
\label{fig:analysis}
\end{figure*}

\noindent\textbf{Improvements on Mask R-CNN~\cite{mask_rcnn}.} Table~\ref{expr:final_results_on_coco} shows the comparison of BN$^*$ (frozen BN), GN and BN$^*$+GCT (using GCT before all the convolutional layers of the backbones). First, we use a short training schedule to compare baselines and GCT counterparts. GCT shows stable and significant improvement in both ResNet-50 and ResNet-101. In ResNet-50, GCT improves box AP by $\mathbf{2.0}$ and mask AP by $\mathbf{1.8}$, which significantly outperforms SE ($0.4$ box AP / $0.5$ mask AP). Moreover, in ResNet-101, GCT also improves detection AP by $\mathbf{1.9}$ and segmentation AP by $\mathbf{1.6}$. Then, we use the long schedule to compare GN and BN$^*$+GCT. GN is more effective than BN when batch size is small as in this case of detection and segmentation using Mask R-CNN. However, we deploy GCT together with BN into the backbone, and these BN$^*$+GCT counterparts achieve much better performance than GN backbones. Compared to GN in ResNet-101, BN$^*$+GCT improves detection AP by $\mathbf{0.8}$ and segmentation AP by $\mathbf{1.1}$. In particular, ResNet-101 with BN$^*$+GCT trained in the short schedule achieves better segmentation AP ($\mathbf{37.7}$) than the GN counterpart ($37.2$) trained with the long schedule. This GN counterpart also uses GN in the backbone, the box heads, and the FPN.
We also explore to combine GCT with GN by introducing GCT into GN box head. The results show GN+GCT achieves a better performance. It demonstrates the benefits of integrating GCT with GN. We now have shown the effectiveness of GCT in working with both BN and GN.

\begin{table}[!t]
\footnotesize
\begin{center}
\setlength{\tabcolsep}{15pt}
\begin{tabular}{|l|c|c|}
\hline  
  Backbone    &    NL-Net~\cite{nlnet} &   \textbf{+GCT}      \\
\hline
ResNet-50    &  74.6   &  \textbf{75.1}$_{(0.5)}$   \\
\hline
ResNet-101   &  75.7   &  \textbf{76.2}$_{(0.5)}$   \\
\hline  
\end{tabular}
\caption{\textbf{Improvement in top-1 accuracy (\%) over the state-of-the-art method on Kinetics.}}\label{expr:final_results_on_kinetics}\vspace{-2mm}
\end{center}
\end{table}

\subsection{Kinetics}
Our previous experiments demonstrate the effectiveness of GCT on image-related tasks. We now evaluate the generalizability in video understanding task of action recognition on a large scale dataset, Kinetics-400~\cite{kinetics}. We employ the ResNet-50 (3D) and ResNet-101 (3D) as the backbone and apply GCT in the last two convolutional layers in each Res-Block. The backbone networks are pre-trained on ImageNet~\cite{imagenet}.

We compare with the state-of-the-art Non-Local Networks (NL-Net)~\cite{nlnet}. The results show that GCT counterparts consistently improves the recognition accuracy over both the ResNet-50 and ResNet-101 baselines, as shown in Table~\ref{expr:final_results_on_kinetics}.
Because of our limited memory resource, we can \textbf{NOT} apply GCT in all the convolutional layers, which we believe can further improve the performance.

In summary, extensive experiments demonstrate that GCT is effective across a wide range of modern architectures, visual tasks, and datasets.

\subsection{Analysis}\label{sec:analysis}
To analyze the behavior of GCT in different layers, we visualize the distribution of the gating weight ($\bm{\gamma}$) of each GCT layer in ResNet-50 on ImageNet. Further, we sort these distributions according to their layer index in 3D space (Fig.~\ref{fig:analysis1}). The bigger layer index means it is closer to the network output. To make the visualization clearer, we re-scale the vertical $z$ axis with $log(1+z)$, which corresponds to the percentage density of $\bm{\gamma}$. We also calculate the mean and standard deviation (std) of $\bm{\gamma}$ in each layer and show them in a bar chart (Fig.~\ref{fig:analysis2}). As shown in Fig.~\ref{fig:analysis1} and~\ref{fig:analysis2}, the mean of $\bm{\gamma}$ tends to be less than $0$ in the GCT layers far from the network output. Oppositely, in the layers close to the output, the mean tends to be greater than $0$. 

According to Eq.~\ref{adaption2} \& \ref{adaption3}, the adaptation of channel $x_c$ is related to $\hat{s}_c$, which corresponds to the ratio of the weighted $\ell_2$-norm of $x_c$ (\ie, $s_c$) and the average of all the $s_c$. When the gating weight $\gamma_c$ is greater than $0$, the adaptation is positively correlated to $\hat{s}_c$ and increases the variance between $x_c$ and others; When $\gamma_c$ is lower than $0$, the adaptation is negatively correlated and reduces the variance.

\begin{table*}[t]

\begin{minipage}[!t]{1\linewidth}
\centering

\subfloat[Embedding operator.\label{expr:ablation_norm}]{
\centering
\footnotesize
\begin{tabular}{|l|c|c|}
\hline
 Norm  & top-1   &   top-5 \\
\hline
$\ell_\infty$ & 23.1 &  6.7 \\
\hline
$\ell_1$ & 22.8 & 6.3 \\
\hline
$\ell_2$ & \textbf{22.7} & \textbf{6.3} \\
\hline
\end{tabular}}\hspace{3mm}
\subfloat[Normalization operator.\label{expr:ablation_normalization}]{
\centering
\footnotesize
\begin{tabular}{|l|c|c|}
\hline
 Normalization   & top-1   &   top-5 \\
\hline
M \& V & 23.7 &  7.1 \\
\hline
$\ell_1$ & 22.9 & 6.4 \\
\hline
$\ell_2$ & \textbf{22.7} & \textbf{6.3} \\
\hline
\end{tabular}}\hspace{3mm}
\subfloat[Adaptation operator.\label{expr:ablation_adaptation}]{
\centering
\footnotesize
\begin{tabular}{|l|c|c|}
\hline
 Adaptation   & top-1   &   top-5 \\
\hline
$Sigmoid$ & 22.9 &  6.5 \\
\hline
$1+ELU$ & 22.7 & 6.4 \\
\hline
$1+tanh$ & \textbf{22.7} & \textbf{6.3} \\
\hline
\end{tabular}}\hspace{3mm}
\subfloat[Application position.\label{expr:ablation_placement}]{
\centering
\footnotesize
\begin{tabular}{|l|c|c|}
\hline
 Position   & top-1   &   top-5 \\
\hline
after BN & 23.1 &  6.6 \\
\hline
before BN & 23.1 & 6.5 \\
\hline
before Conv & \textbf{22.7} & \textbf{6.3} \\
\hline
\end{tabular}}
\vspace{-2mm}
\caption{\textbf{Ablation experiments.} We evaluate error performance in GCT-ResNet-50 on ImageNet (\%). The ResNet-50 baseline achieves a top-1 of $23.8$ and a top-5 of $7.0$. M \& V denotes the mean and variance normalization.}\label{tab:ablations}
\end{minipage}\vspace{-2mm}
\end{table*}

Based on the analysis and the results we observe, we suppose that GCT tends to reduce the difference among channels in layers far away from the output. This behavior is helpful to encourage cooperation among channels and relieve overfitting. Apart from this, GCT tends to increase the difference among channels when close to the output. Here, GCT acts like attention mechanisms that focus on creating competition.

To further validate our hypothesis, we calculate the ratio of the variance of output and input feature of each GCT layer, which we show in Fig.~\ref{fig:analysis3}. 
Generally, the shallow convolutional layers learn low-level attributes to capture general
characteristics like textures, edges, and corners. Here, GCTs reduce the feature variances to avoid missing some attributes. Besides, the feature variances become larger in deeper layers, where the high-level features are more discriminative and task-related.
As expected, in the layers close to network output, GCT tends to magnify the variance of input feature (the ratio is always greater than $1$), but in the layers far away from the output, GCT tends to reduce the variance (the ratio is always less than $1$). This phenomenon is consistent with our previous hypothesis and shows that GCT is effective in creating both competition and cooperation among channels.
Our observation validates that GCT can adaptively learn the channel relationships at different layers.

As shown, for the stages far away from the network output, the proposed GCT layer tends to reduce the variance of input feature, which encourages cooperation among channels and avoids excessive activation values or loss of useful features.
On the contrary, for those stages close to the output, GCT tends to magnify the variance. These phenomena are consistent with the observation in SE and its following work~\cite{eca}, \ie, the attention weights are shared in shallower layers, but more discriminative in deeper layers.

\subsection{Ablation Studies}
In this section, we conduct a serial of ablation experiments to explain the relative importance of each operator in the GCT. At last, we show how the performance changes with regards to the GCT position in a network.

\noindent\textbf{Embedding component.}
\label{sec:embedding}
To explain the importance of $\ell_p$-norm in the embedding module, we compare embedding operators with different $\ell_p$ norm. We report the results in Table~\ref{expr:ablation_norm}, which shows all the $\ell_p$-norms are effective, but the $\ell_2$-norm is slightly better than $\ell_1$-norm. Besides, we make a clock time comparison between SE and GCT with different embedding components. As shown in Table~\ref{tab:clock}, $\ell_2$-norm is computationally similar to $\ell_1$-norm and GCT is much more efficient than SE. The results demonstrate that $\ell_p$-norms are robust and perform better.

\noindent\textbf{Normalization component.}
\label{sec:normalization}
We also explore the significance of $\ell_p$-norm in the channel normalization by comparing $\ell_p$ normalization with mean and variance normalization. The mean and variance normalization will normalize mean to $0$ and variance to $1$, which is widely used in normalization layers (\eg,~\cite{bn}). We show all these results in Table~\ref{expr:ablation_normalization}. Particularly, mean and variance normalization achieves a top-1 error of $23.7\%$, which is only slightly better than the ResNet-50 baseline ($23.8\%$). Both $\ell_1$ and $\ell_2$ normalization make more promising improvements, and $\ell_2$ performs slightly better. 

$\ell_p$ normalization is better at representation learning in channel normalization.

\noindent\textbf{Adaptation component.}
\label{sec:adaptation}
We replace the activation function of the gating adaptation with a few different non-linear activation functions and show the results in Table~\ref{expr:ablation_adaptation}. Compare to the baseline (top-1 of $23.8\%$), all the non-linear adaptation operator achieves promising performance, and $1+tanh$ achieves a slightly better improvement. Both $1+tanh$ and $1+$ELU~\cite{elu}, which employ residual connection and can model identity mapping, achieve better results than $Sigmoid$. \Zongxin{These results show that the residual connection is important for the gating adaptation.}

\noindent\textbf{Application position.}
\label{sec:appication}
 To find the best way to deploy GCT layers, we conduct experiments in ResNet-50 architecture on ImageNet by separately applying GCT after all the BN layers, before all the BN layers, and before all the convolutional layers. The results are reported in Table~\ref{expr:ablation_placement}. All the placement methods are effective in using GCT to improve the representational power of networks. However, it is better to employ GCT before all the convolutional layers, which is similar to the strategy in~\cite{alexnet} (normalization after ReLU).

\section{Conclusion and Future Work}

In this paper, we propose GCT, a novel layer that effectively improves the discriminability of deep CNNs by leveraging the relationship among channels. Benefit from the design of combining normalization and gating mechanisms, GCT can facilitate two types of neuron relations, \ie, competition and cooperation, with negligible complexity of parameters. We conduct extensive experiments to show the effectiveness and robustness of GCT across a wide range of modern CNNs and datasets.
Except for CNNs, recurrent networks (\eg, LSTM~\cite{lstm}) are also a popular branch in deep neural networks area. 
In future work, we will study the feasibility of applying GCT into recurrent networks.

\noindent\textbf{Acknowledgements.} This work is supported by ARC DP200100938.

{\small
\bibliographystyle{ieee_fullname}
\bibliography{egbib}
}

\end{document}